# Robot Operator: Social Robot Based on Large Language Model and Machine Learning for Real-time Phone Fraud Prevention

Chen-jun Liu [1,2], Bo-wei Hsu [1], Yi-cheng Liu [1], Guan-ru Lai [3], An-Te Tsai [4], Lyn Chao-ling Chen (Corresponding) [5,*]
Department of Computer Science, National Chengchi University, Taipei, Taiwan [1],
Bachelor Program in Artificial Intelligence Applications, Taipei, Taiwan [2],
Graduate Institute of Artificial Intelligence Cross-disciplinary Technology, National Taiwan University of Science and Technology, Taipei, Taiwan [3],
Department of Statics, National Taiwan University of Science and Technology, Taipei, Taiwan [4],
Interdisciplinary Artificial Intelligence Center, National Chengchi University, Taipei, Taiwan [5],
112703016@g.nccu.edu.tw [1,2], 115753111@g.nccu.edu.tw [1], louisliu911231@gmail.com [1],
M11552023@mail.ntust.edu.tw [3], 110405193@g.nccu.edu.tw [4], lynchen@ntu.edu.tw [5,*]

## ABSTRACT

**In the work, the real-time fraud prevention was considered in the study, and the humanoid robot acts as an operator to answer the VoIP phone call in home scenario. In the training phase, the K-Means machine learning method was adopted to classify 1 safe stage and 3 fraud stages of phone fraud using the real dataset. GPT-4o large language model was adopted for the phone agent, and a web server was built to transfer the VoIP data between the phone agent and the social robot. Once a VoIP phone call is coming, the phone agent answers the phone call automatically, and guide the caller to clam the calling purpose. In the time period of Speech-to-Text conversion, the phone agent triggers a voice message via the robot when the semantic analysis results match the fraud stages; otherwise, the phone agent send a safe message showing on the screen of the robot to notify the users a safe phone call in real-time. The contribution of the work that integrates the artificial intelligence technologies and the socially assistive robot for real-time phone fraud prevention, and the robust of the system that can be applied in real scenario.**

## I. INTRODUCTION

Fraud has become a serious problem in Taiwan society, over half of people (24,731 people across 13 regions) in Taiwan encounter scams each week, and almost 30% of victims hand over money or personal data within the first hour from the Asia scam report in Taiwan of GASA, ScamAdviser, and Gogolook (Whoscall) [1]. From the report, phone fraud is the major scam type (59.9%), the other scam types are social media (54.5%) and Short Message Service (SMS) (52.5%) [1]. Scammers use similar communication phrases in phone fraud such as “You have an overdue..,” “I can help you..,” or “Let’s settle this together,” and people often aware the risk of phone fraud, however, trust and panic are the main reasons cause the victims [2]. Hence, in the work, the Socially Assistive Robot (SAR) (NUWA Kebbi Air S) (Figure) integrates Artificial Intelligence (AI) technologies for real-time phone fraud prevention.

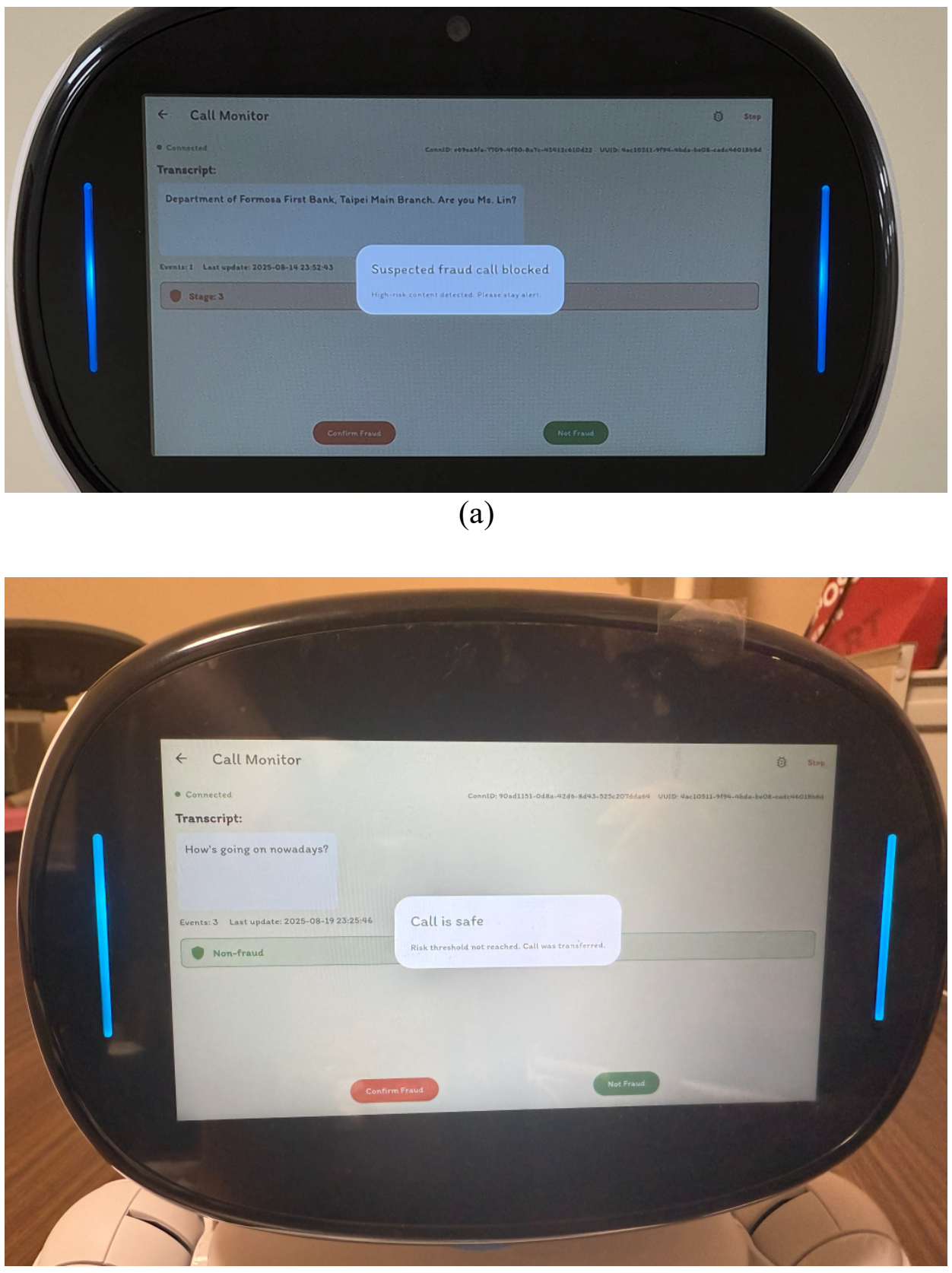

(a)

(b)

Figure 1. Real-time phone fraud prevention: (a) scam phone call and (b) safe phone call.

This paper is structured as follows: Section 2 reviews the researches in phone fraud detection. Section 3 outlines the methodology that introduces the data preprocessing, phone fraud stage classification and Application (APP) of the SAR. Section 4 concludes the study.

## II. RELATED WORKS

In the aspect of scam script analysis in Natural Language Processing (NLP) field, using the data from the community of "scam baiters'" on YouTube, topic and time series modeling and emotion recognition were adopted to recognize to scammer utterances that the scammer used emotions as a social engineering tool obviously [3]. Anti-Social Engineering Tool (ASsET) was proposed to detect phone frauds based on the semantic content of the conversation, and word embedding techniques was adopted to determine if the meaning of a scam signature is contained in a conversation [4]. From the public dataset of stories of victims in Singapore, Term Frequency-Inverse Document Frequency (TF-IDF) was used to extract key terms in scam stories, and the text preprocessing includes removal of Unicode characters, digits, and contracted words, conversion to lower cases, removal of spaces and punctation marks, unabbreviation of acronyms, and correction of misspelled words [5]. After translation the audio files of phone calls through Google Cloud Speech-to-Text (STT) Application Programming Interface (API), NLP stemming was adopted to extract words from the phone fraud dataset that the text preprocessing includes removing English stopwords, punctuations, symbols, and numbers, and machine learning method was adopted for scam call classification [6]. Retrieval Augmented Generation (RAG) was adopted in real-time phone fraud detection that the callers solicit private information of users [7]. Hence, phone crime script was considered in the work.

In the aspect of adopting agent in fraud prevention, RoboHalt combined multiple lightweight NLP/Machine learning (ML) sub-models to distinguish real phone call from human and the machine calls, and acts as a middleware on cell phone or within network carrier that only transferred the real phone calls [8]. Hence, in the work, the SAR acts as an operator to answer the phone call automatically, and only transfers the safe calls to the users.

## III. METHODOLOGY

### A. Data Augmentation Through Generative Artificial Intelligence in Phone Fraud Keyword Table

In the preprocessing phase, the dataset was collected from the real fraud articles on the internet forum such as Quora and Reddit, and correct the sentences by adding or adjusting commas manually. After calculating of TF-IDF scores, a keyword table (Table I) was built from the top 10% of high-scoring (Threshold>4.0), and only nouns were considered eliminating the non-informative words and the stop words (Implemented in Python), and 5 synonyms of each keyword are added manually (Total of 6 terms). In the phase of data augmentation, for improving the accuracy of the input of sentences in the stage classification, the Generative Pre-trained Transformer 4 Omni (GPT-4o) model was adopted for generating synonyms of the original keywords. The token positions of the keywords in the sentence are also marked for generating difference composition of the original sentence with semantic similarity, and the prompt is as follows: "*You are a data-augmentation bot.\n Replace ONLY the tokens listed below with suitable synonyms of the same. Part-of-speech. Keep order and punctuation.\n Return exactly {n} sentences as a JSON array, without the original.*" The examples of original sentence such as "Not asking for money but to contact this person," and the generated sentence such as "Not asking for cash but to contact this person" or "Not asking for currency but to contact this person." Hence, the augmented keyword table is used to determine the potential fraud, and the sentences with matched keywords will continue process in the stage classification.

TABLE I. PHOBE FRAUD KEYWORDS IN REAL DATA

| **Original Keyword (Non) (Order by TF-IDF from the highest)** | **Synonyms of Original Keywords** |
|---|---|
| account | report, page, profile, portal, record |
| accounts | profiles, records, logins, user accounts, ledgers |
| bank | finance, institution, branch, lender, trust |
| card | badge, pass, token, credit card, voucher |
| case | episode, event, matter, instance, issue |
| company | corporation, organisation, firm, business, enterprise |
| credit | loan, funds, score, rating, installment |
| details | particulars, information, specifications, data, facts |
| email | mail, message, e-mail, note, inbox |
| id | identifier, ID number, license, code, tag |
| information | details, data, facts, intel, info |
| message | note, notification, text, mail, alert |
| money | funds, cash, capital, finance, currency |
| office | bureau, department, agency, office, branch |
| order | purchase, request, command, order, booking |
| police | authorities, law enforcement, officers, law, constables |
| problem | issue, difficulty, trouble, error, fault |
| security | protection, safety, guard, defence, safekeeping |
| support | assistance, help, service, aid, succor |
| virus | malware, bug, infection, trojan, worm |
| wife | spouse, partner, consort, missus, mate |

### B. Machine Leaning in Phone Fraud Stage Classification

From the dataset of real fraud articles, only the sentences with terms (Keyword: $\geq$1) in the keyword table are vectorized (1217 sentences). From the distributions of the results, word position in the sentences without obvious features in the phone fraud stage (Figure 2 and 3). After dimensionally reduction through Principal Component Analysis (PCA), classification, machine learning of K-Means clustering was adopted to classify the sentences into stages according to the semantic similarity,

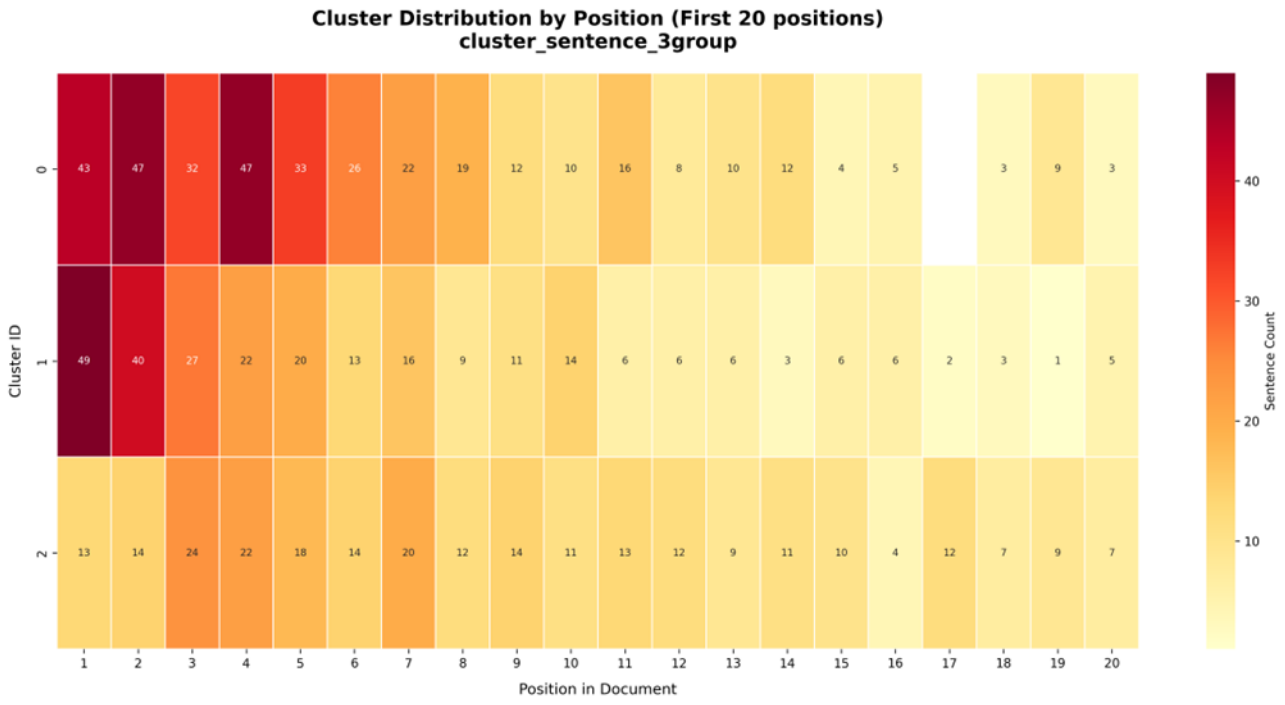

(a)

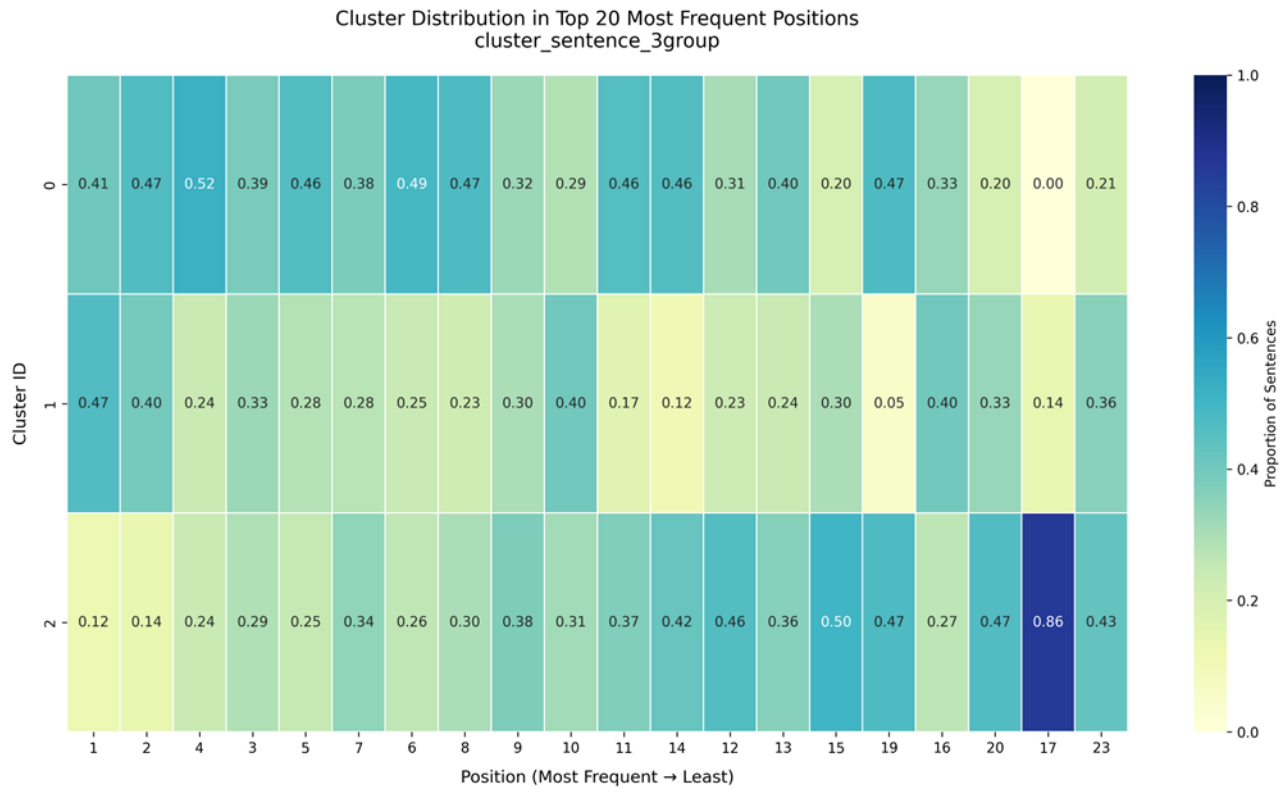

(a)

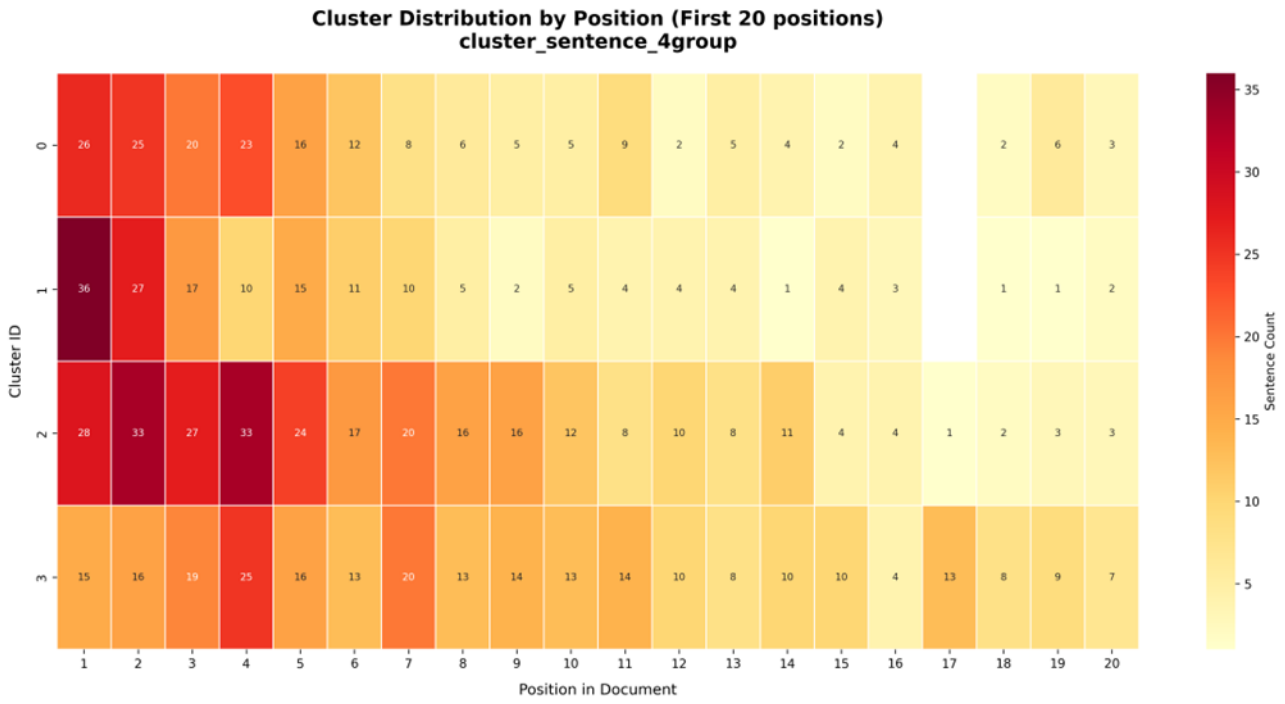

(b)

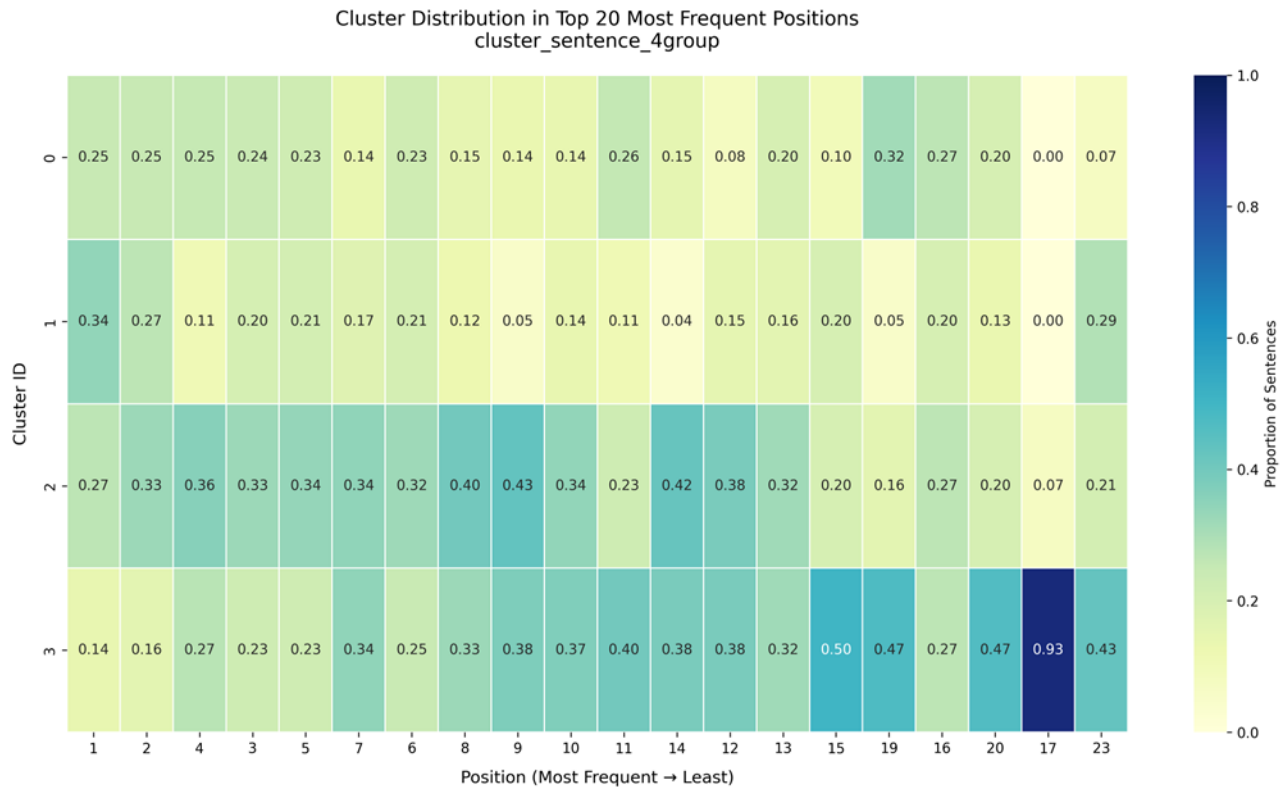

(b)

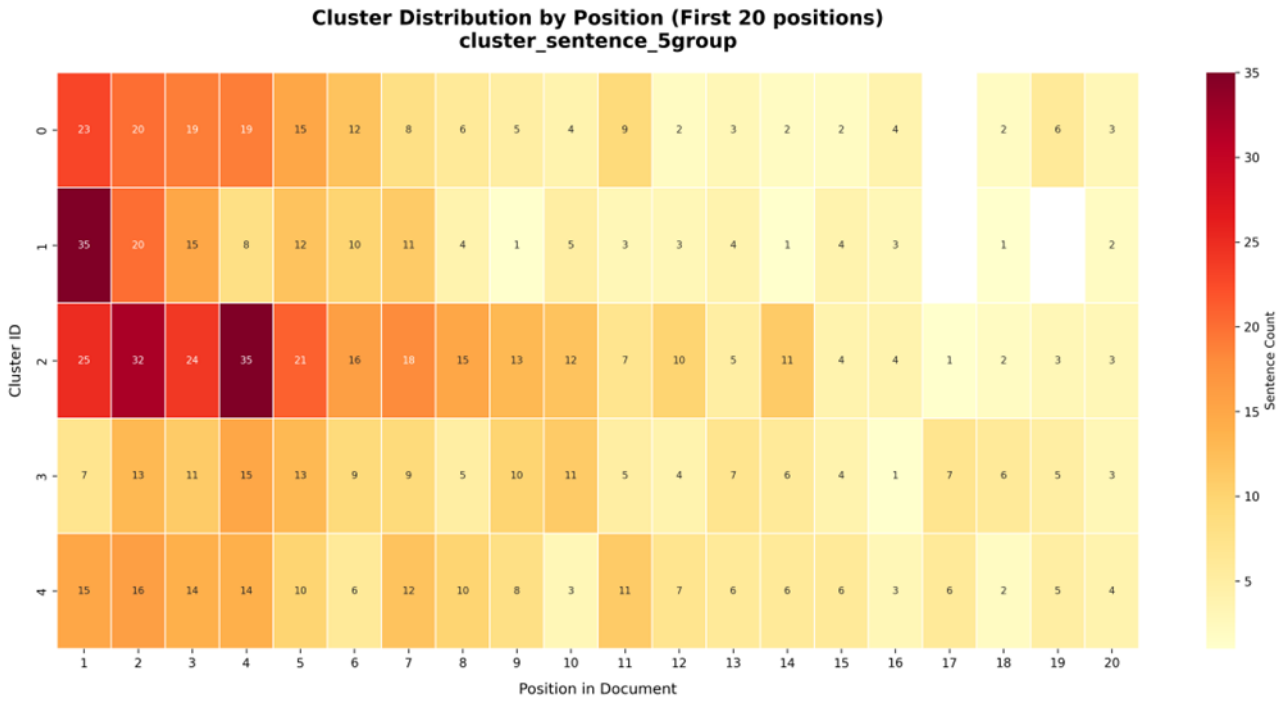

(c)

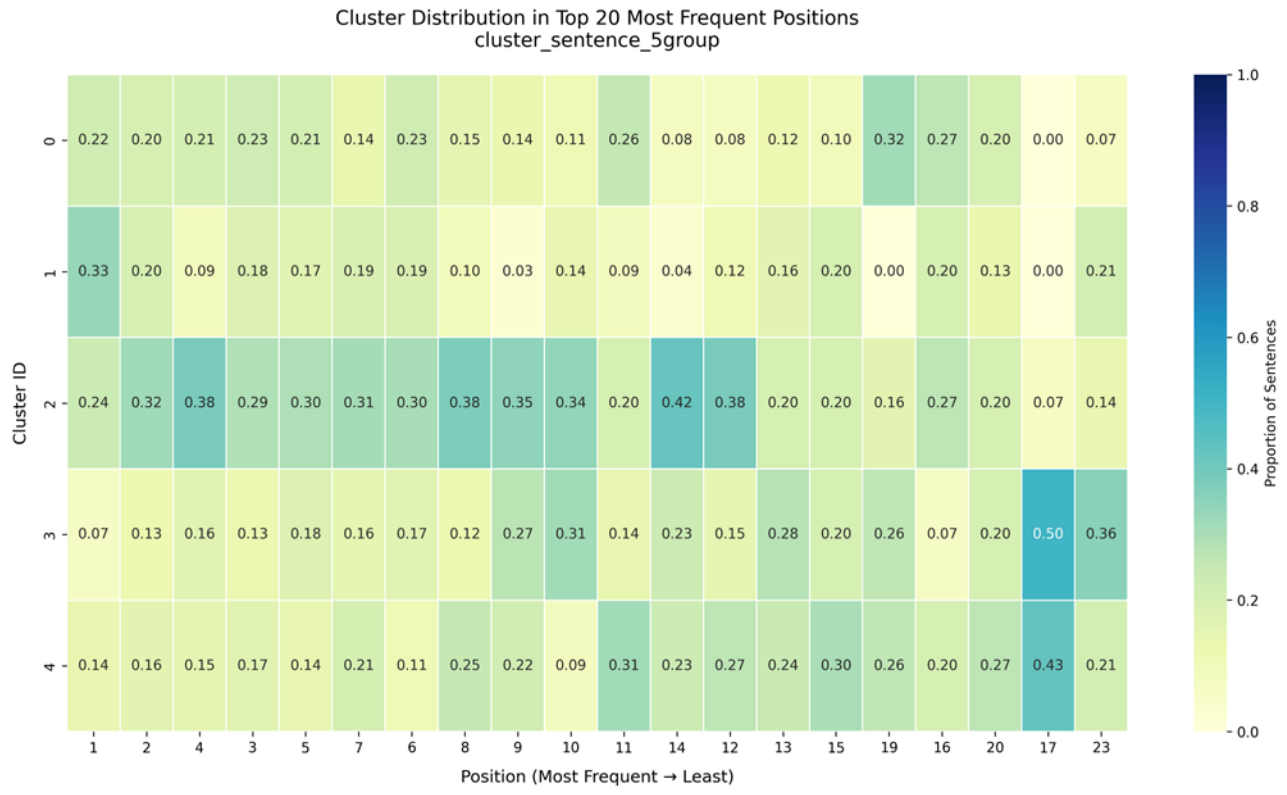

(c)

Figure 2. Position in cluster diverge: (a) 3 groups, (b) 4 groups and (c) 5 groups.

Figure 3. Frequency of position: (a) 3 groups, (b) 4 groups and (c) 5 groups.

and the results show the best classification in 4 stages (K=4) with distinct semantic features (Figure 4). In addition, the After classification, the example sentences in the stage 0, 1, 2, 3 such as "Its important to never click links in emails unless absolutely necessary, for example a password reset link you requested or an account activation link for an account you created," "Remember the scam, where the caller claim to be from the IRS, telling that your account had done a bad job and the there was a warrant out for your arrest at 3pm this afternoon," "ma'am I'm talking about did you provide any of your personal information in the public like your friends family members," "It looks like they dont need my password you can just get Amazon to send me this OTP message to any phone number by just entering my phone number and sign in with passcode," respectively. According to the classification results, stage 0: fraud verification (240 sentences) has defined as a safe Voice over Internet Protocol (VoIP) phone call, and stage 1: fraud experience sharing (248 sentences), stage 2: internet fraud reminder (342 sentences) and stage 3: technology support fraud (387 sentences) has defined as a fraud VoIP phone call (Implemented in Python). Hence, the stage classification determines the phone call is a safe or a fraud VoIP phone call.

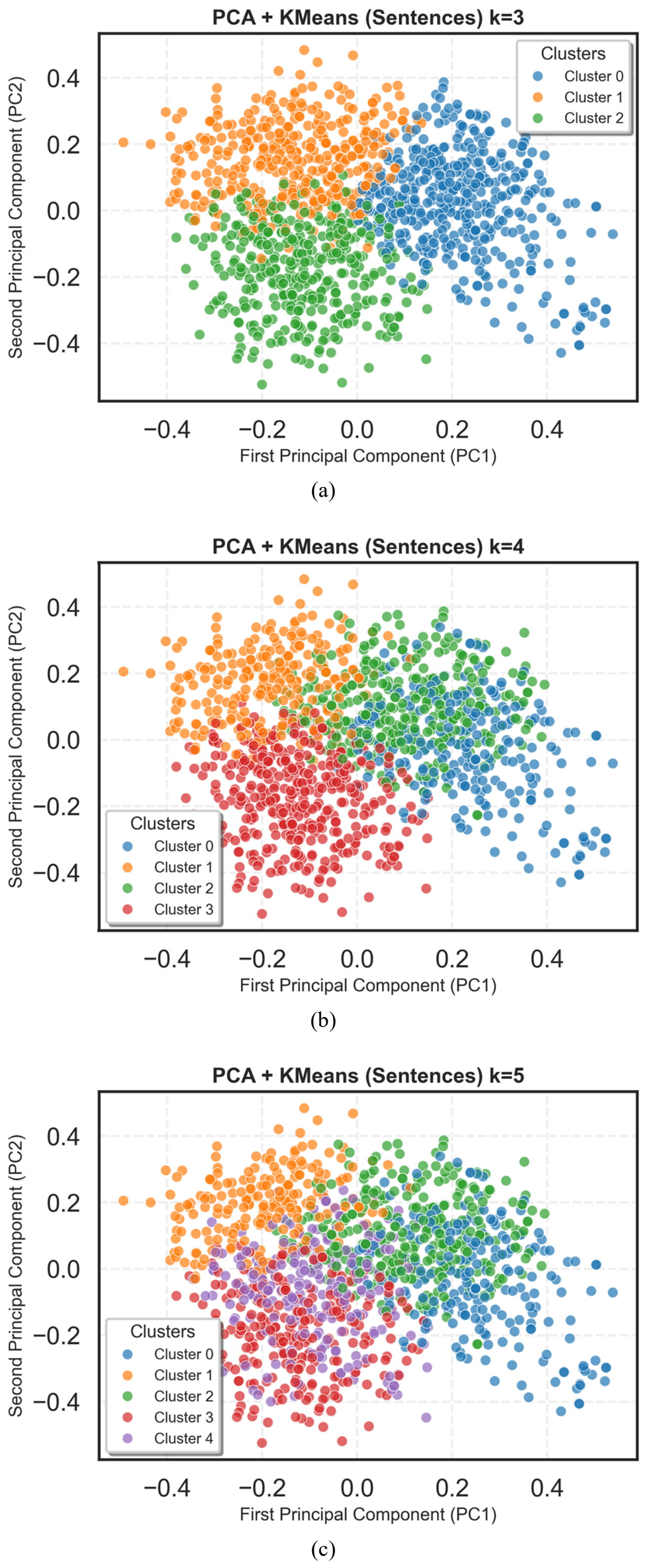

Figure 4. PCA Visualization of K-Means Clusters on sentences embeddings: (a) K=3, (b) K=4 and (c) K=5.

### C. *Real-time Warning Application in Socially Assistive Robot*

In the interface design of SAR warning, users start the APP on the SAR (Figure 5 (a)), and allows the APP to record the VoIP phone call (Figure 5 (b)), and the APP monitors the incoming VoIP phone continuously (Figure 5 (c)). Faster Whisper model was adopted to covert audios signals to texts of a VoIP phone call, and links the VoIP phone call to the robot (Implemented in Google Flutter SDK, Dart, Kotlin and NuwaRobot SDK) [9]. Asynchronous Server Gateway Interface (ASGI) web server was built to make a VoIP phone call and to transfer VoIP phone call data to the robot, and the database has built though PostgreSQL (Implemented in Python FastAPI and Uvicorn). WebSocket and Web Real-Time Communication (WebRTC) were both used to transfer voice data between the webpage and the robot for robust consideration. In addition, Secure Sockets Layer SSL certification (Let's Encrypt) and Nginx reverse proxy provide secure VoIP phone call.

For preventing the users to answer a fraud VoIP phone call directly, the GPT-4o model was adopted for generating voice messages to reply a VoIP phone call, and the prompt input is as follows for guiding the caller to explain the calling purpose: *"Imagine you are a person having a conversation with scammers, and you are playing with them, trying to gather information while keeping the caller engaged. Your primary objectives are: 1. Engage the caller in conversation to stall for time without revealing any personal, financial, or sensitive information. 2. Use polite but non-committal responses to keep the scammer talking while avoiding giving them useful data. 3. Subtly prompt the caller to reveal more about their intentions, methods, or identity, if possible. 4. Maintain a calm, cooperative, and human-like tone, without making the caller suspicious that you are an AI or that you are intentionally stalling. 5. Under no circumstances should you provide real personal information or agree to perform any actions requested by the caller. 6. Your ultimate goal is to protect yourselves while gathering as much information as possible from the scammer, delaying them and reducing the risk of successful fraud."* [10].

Given an input of audio signals in a time period (3 minutes), the sentences are filtered using a keyword table after GenAI data augmentation with defined fraud terms and the correlated synonyms. If sentences match the keyword table were identified as a possible fraud VoIP phone call, the sentences are further verified with the predefined fraud stages through semantic analysis to classify into 4 stages (Stage 0: fraud verification, stage 1: fraud experience sharing, stage 2: internet fraud reminder, stage 3: technology support fraud). Half of sentences in the time period in the fraud stage (Stage: 1, 2, 3) are considered as a fraud VoIP phone call, triggers the warning voice messages "Fraud warning" (3 times) of the robot, and showing the texts of the voice messages from the caller and a notification of "Suspected fraud call blocked" on the screen of the robot (Figure 1 (a)). Meanwhile, the Large Language Model (LLM) reply to the VoIP phone call continuously. Otherwise, the robot receives a notification via web socket from the server in the safe stage (Stage: 0), and notifies the users to answer the VoIP phone call, showing the texts of the voice messages from the caller and a notification of "Call is safe" on the screen of the robot (Figure 1 (b)). Hence, the real-time fraud VoIP phone call detection has realized.

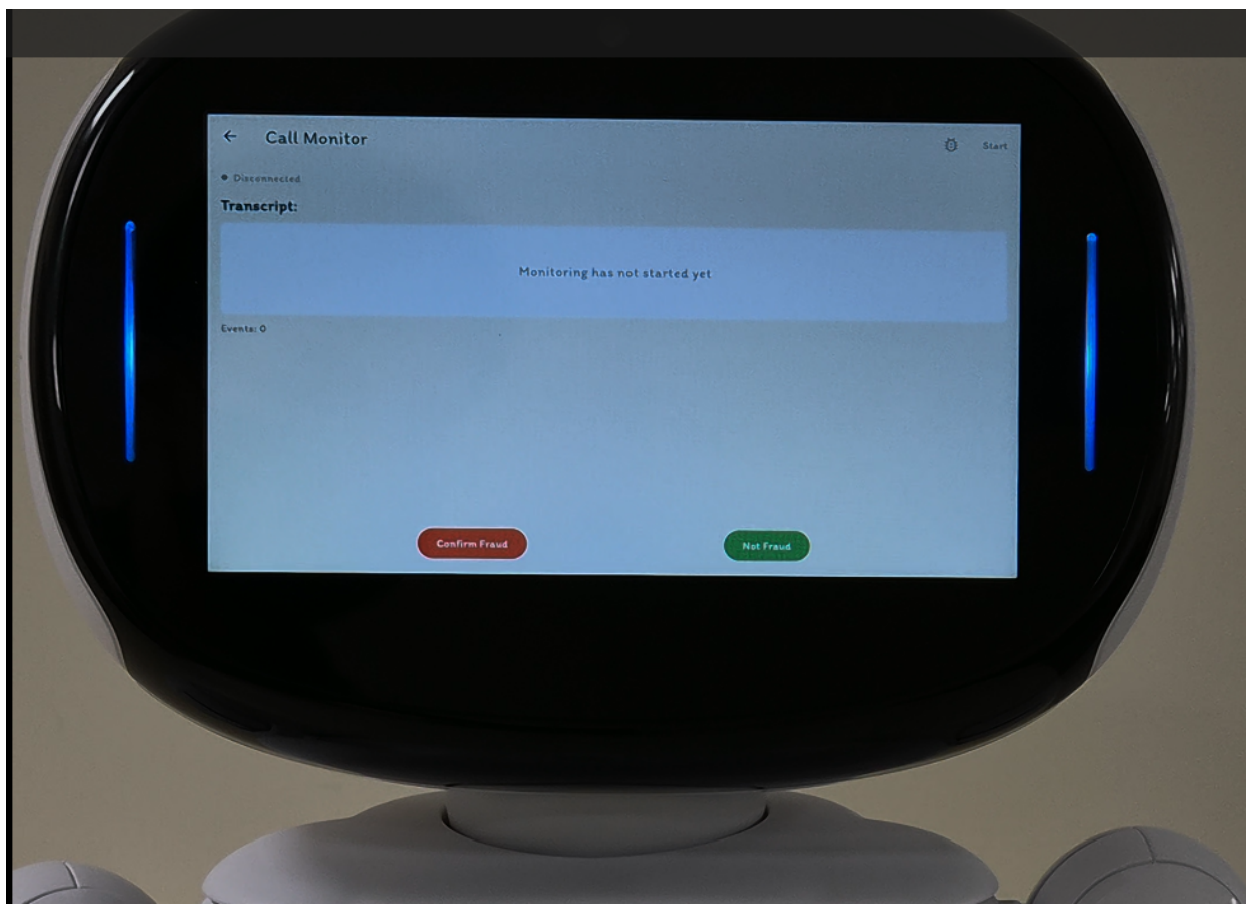

(a)

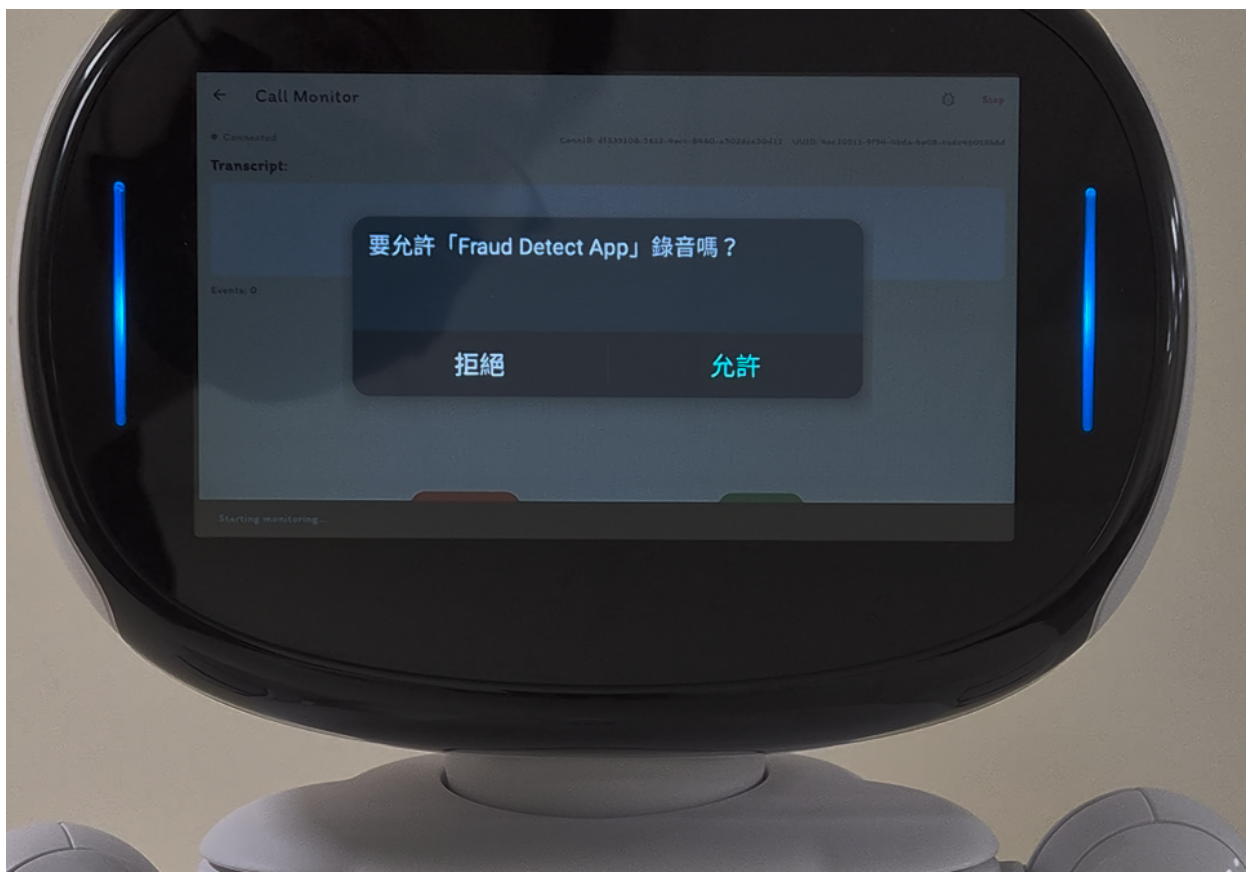

(b)

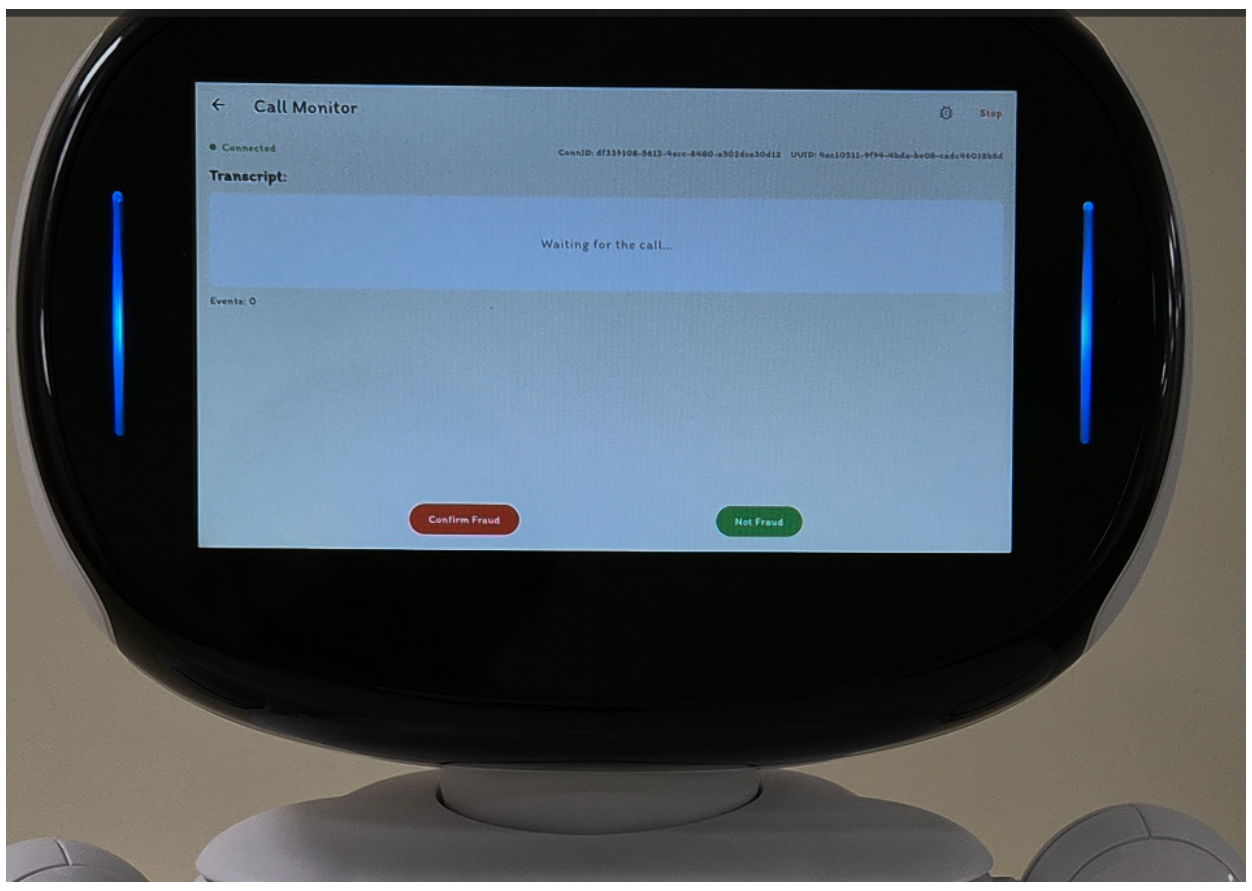

(c)

Figure 5. APP in the SAR: (a) initial, (b) user permission and (c) phone fraud monitoring.

## IV. CONCLUSION

A real-time phone fraud prevention APP using the SAR has developed, and the SAR acts as an operator to answer the phone call automatically in home scenario. In addition, the SAR have scalability by improving the performance of phone fraud detection model for facing the fast changes of phone frauds.

## ACKNOWLEDGMENTS

This work was partially supported by the National Science and Technology Council, Taiwan, under grants 114-2635-E-004-001-. Special thanks Shih-chan Chiu, Min-jui Sung, Li-chen Ko and Yu-cheng Chen for the assistance with the literature research.